\documentclass[preprint]{article}

\usepackage{style}

\usepackage{natbib}
\setcitestyle{numbers}
\usepackage{amsmath,amsfonts,amssymb}
\usepackage[utf8]{inputenc} 
\usepackage[T1]{fontenc}    
\usepackage{hyperref}       
\usepackage{cleveref}
\usepackage{url}            
\usepackage{booktabs}       
\usepackage{nicefrac}       
\usepackage{microtype}      
\usepackage{xcolor}         
\usepackage{graphicx}
\usepackage{glossaries}
\usepackage{subcaption}
\usepackage{tikzducks}
\usepackage{siunitx}
\usepackage{subcaption}

\usepackage{comment}
\usepackage{todonotes}

\newcommand{\onet}{\hat{w}_\ell^{\text{ONet}}}
\newcommand{\nbf}{\hat{w}_\ell^{\text{NBF}}}

\DeclareFontFamily{U}{matha}{\hyphenchar\font45}
\DeclareFontShape{U}{matha}{m}{n}{
      <5> <6> <7> <8> <9> <10> gen * matha
      <10.95> matha10 <12> <14.4> <17.28> <20.74> <24.88> matha12
      }{}
\DeclareSymbolFont{matha}{U}{matha}{m}{n}
\DeclareMathSymbol{\odiv}         {2}{matha}{"63}

\glsdisablehyper
\newacronym{AoA}{AoA}{angle-of-attack}
\newacronym{AD}{AD}{automatic differentation}

\newacronym{BC}{BC}{boundary condition}

\newacronym{CFD}{CFD}{computational fluid dynamics}

\newacronym{ONet}{DeepONet}{deep operator network}

\newacronym{FEM}{FEM}{finite element method}
\newacronym{FNO}{FNO}{Fourier neural operator}

\newacronym{ML}{ML}{machine learning}
\newacronym{MLP}{MLP}{multi-layer perceptron}
\newacronym{MSE}{MSE}{mean squared error}

\newacronym{NBF}{NBF}{neural basis function}
\newacronym{NN}{NN}{neural network}
\newacronym{NIG}{NIG}{normal inverse-gamma}
\newacronym{NLL}{NLL}{negative log-likelihood}
\newacronym{NSE}{NSE}{Navier-Stokes equations}

\newacronym{PDE}{PDE}{partial differential equation}
\newacronym{PINN}{PINN}{physics-informed neural network}
\newacronym{PINO}{PINO}{physics-informed neural operator}
\newacronym{POD}{POD}{principal orthogonal decomposition}
\newacronym{PODONet}{principal orthogonal decomposition deep operator network}{POD-ONet}

\newacronym{RAM}{RAM}{Radio Attenuation Measurement}
\newacronym{ROM}{ROM}{reduced order modeling}

\newacronym{SciML}{SciML}{scientific machine learning}
\newacronym{SGD}{SGD}{stochastic gradient descent}
\newacronym{SVD}{SVD}{singular value decomposition}

\newacronym{UQ}{UQ}{uncertainty quantification}

\title{Data-efficient operator learning for solving high Mach number fluid flow problems}
\author{Noah Ford$^1$,\,\,\,\,Victor J. Leon$^2$,\,\,\,\,Honest Mrema$^3$,\,\,\,\,Jeffrey Gilbert$^3$,\,\,\,\,Alexander New$^2$\\
    $^1$ Force Projection Sector \\
    $^2$ Research and Exploratory Development Department \\
    $^3$ Air and Missile Defense Sector \\
    Johns Hopkins University Applied Physics Laboratory\\
    Laurel, Maryland 21044\\
    \texttt{\{noah.ford, victor.leon, honest.mrema, jeffrey.gilbert, alex.new\}@jhuapl.edu}
}

\begin{document}

\maketitle
\begin{abstract}
    We consider the problem of using \gls{SciML} to predict solutions of high Mach fluid flows over irregular geometries. In this setting, data is limited, and so it is desirable for models to perform well in the low-data setting. We show that the \gls{NBF}, which learns a basis of behavior modes from the data and then uses this basis to make predictions, is more effective than a basis-unaware baseline model. In addition, we identify continuing challenges in the space of predicting solutions for this type of problem.
\end{abstract}

\glsresetall

\section{Introduction}\label{sec:introduction}


Scientific models enable scientists and engineers to study complex systems of interest, such as animal brains~\cite{Vendel2019drugs}, bacteria growth~\cite{Ford2020bacteria}, and fluid dynamics~\cite{jin2021nsfnets}. 
These models are typically computationally intensive, which limits how much they can be used.
An increasingly-common alternative is to supplement these models with \gls{SciML} approaches~\cite{karniadakis2021physics} that incorporate an inductive bias to improve data efficiency.
Examples of physics-informed inductive biases include the physics-informed \gls{ONet}~\cite{Wang2021onet,Wang2023onet} and \gls{PINO}~\cite{Li2021pino}, or the \gls{PODONet}~\cite{Lu2022poddeeponet}. Another type of inductive bias is the use of basis functions to form the model predictions, which limits the model's expressiveness. In this paper, we explore use of basis functions and this bias' effect on predictive performance. 


In this work, we analyze high-Mach fluid flow over a blunt nose cone (\Cref{fig:cone}, with further details in~\Cref{sec:data_generation}). Compared to other \gls{SciML} settings: (1) There are no existing large-scale databases for its behavior, (2) it has a non-uniform geometry, and (3) its high-Mach fluid flows create complex physical features like boundary layers. Thus, we rely on an implementation of the recent \gls{NBF}~\cite{Witman2022nbf} as our primary \gls{SciML} model, which has been applied to similar problems before, while also comparing to a \gls{ONet}~\cite{Lu2021deeponet}-like architecture that lacks scientific regularization. In this study, we do not use scientific regularization when fitting the \gls{NBF}. By omitting scientific regularization, we increase the \gls{NBF}'s similarity to the physics-unaware \gls{ONet}, improving the comparability of the two models.

The \gls{NBF} combines the \gls{ONet} with ideas from \gls{ROM}, in particular the \gls{POD}~\cite{sirovich1987turbulence}, to first learn a basis representation for the system's state variables and then an additional set of functions that combine the basis representation into variable predictions. The basis acts as a form of regularization that helps deal with complex physical features. Similar techniques have been used, e.g., the \gls{POD}-\gls{ONet}~\cite{Lu2022poddeeponet}.

Compared to a basis-unaware \gls{ONet}, the basis regularization of the \gls{NBF} improves the relative prediction error in the low-data regime when predicting solutions to high Mach fluid flow problems (\Cref{tab:eval_results}). However, both models are unable to fully fit the training data (\Cref{tab:training_results}). Additionally, accurate prediction of the density variable $\rho$ is difficult, due to its highly skewed distribution. This motivates the need for further development in expressive neural architectures and training schemes that can learn complex distributions of \gls{PDE} behavior.



\section{Methods}\label{sec:methods}

\subsection{Problem setup}\label{sec:problem_setup}

We consider a dataset $\mathcal{D}$ that consists of sets $(X, W^d, \psi^d),d=1,\hdots,D$, where $X = \{x_n\}_{n=1}^N\subseteq\Omega\subseteq\mathbb{R}^{N_x}$ is a spatial mesh over an irregular geometry $\Omega$, shared across all  sets, $W^d = \{w^d_n\}_{n=1}^N\subseteq\mathbb{R}^{N_w}$ is a set of state variables values (with $w^d_n$ the value at mesh point $x^d_n$), and $\psi^d\in\mathbb{R}^{N_\psi}$ is a parameter vector. We learn  models $\hat{w}_\ell$, one for each state variable $\ell=1,\hdots,N_w$. 

\begin{figure}
    \centering
    \begin{subfigure}{0.45\linewidth}
        \includegraphics[width=0.6\linewidth]{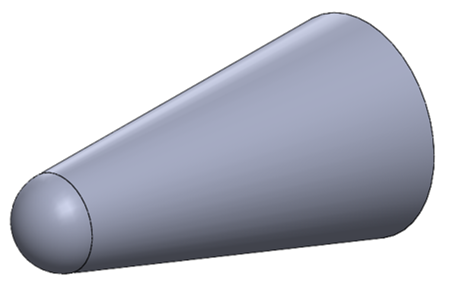}
        \caption{The RAM-C II blunt nose cone}
        \label{fig:cone}
    \end{subfigure}
    \qquad
    \begin{subfigure}{0.45\linewidth}
        \includegraphics[width=0.6\linewidth]{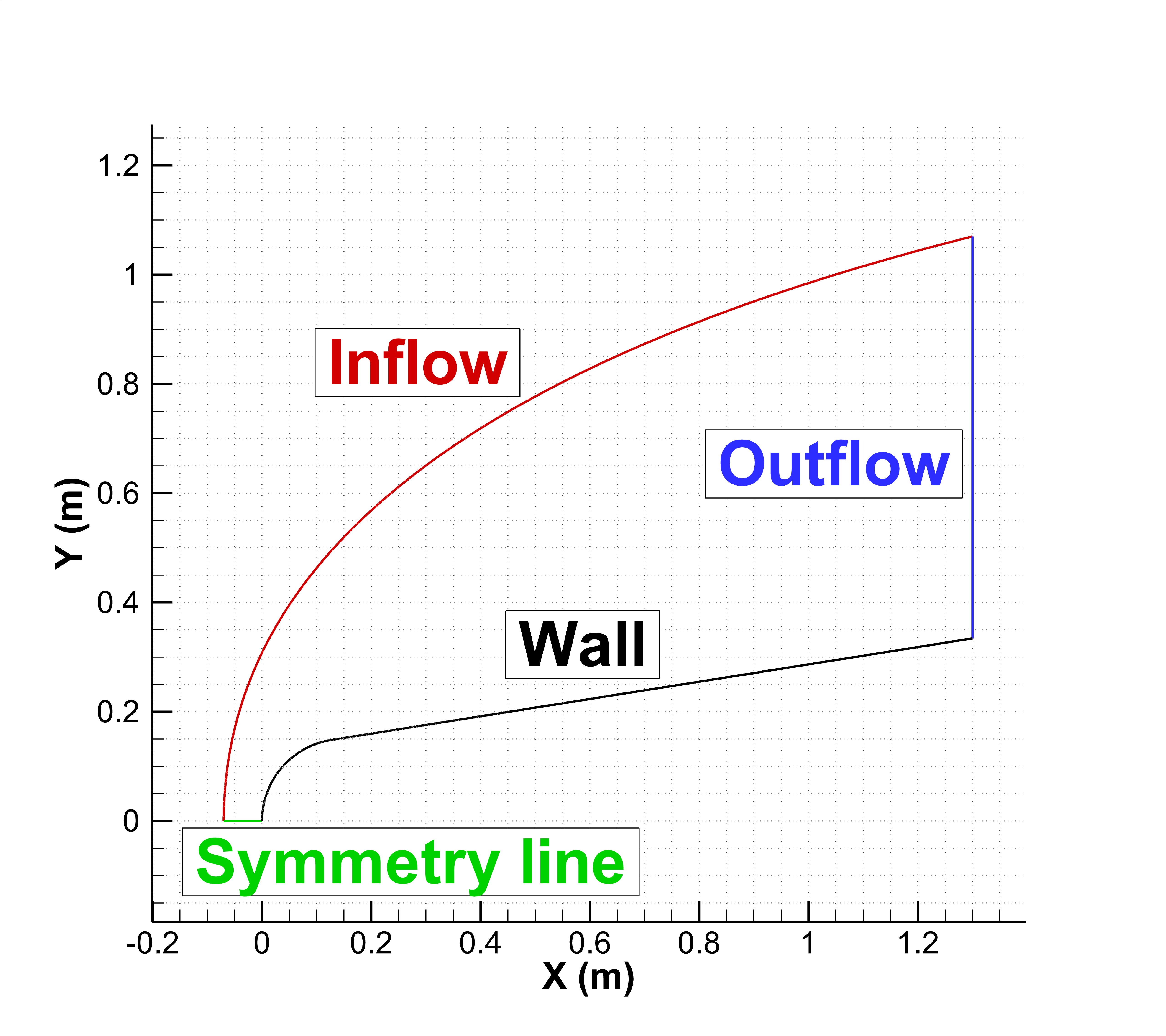}
        \label{fig:mesh}
        \caption{The domain components}
    \end{subfigure}
    \caption{The RAM-C II geometry}
    \label{fig:ram-cII}
\end{figure}

In our setting (\Cref{sec:data_generation}), data satisfy compressible \gls{NSE} (\Cref{sec:governing_equations}) defined over a 3D axisymmetric geometry (\Cref{fig:ram-cII}), based on the \gls{RAM}-C II flight vehicle~\cite{Farbar20213ramcII,Sawicki2021ramcII}. Due to the axisymmetry, there are two spatial inputs ($N_x=2$). We model four state variables: $x$-velocity ($u_1$), $y$-velocity ($u_2$), density ($\rho$), and temperature ($T$). The parameter vector $\psi$ has two components: Mach and altitude. Each solution is resolved on a mesh of $N=160250$ spatial points.

\subsection{ML models for predicting high Mach fluid flow}\label{sec:ml_models}

To solve the RAM-C II fluid flow problem, we rely on two types of \gls{SciML} model. The first is a physics-unaware \gls{ONet}~\cite{Lu2021deeponet}, and the second is an implementation of the \gls{NBF}~\cite{Witman2022nbf}.

For a state variable $\ell$, the \gls{ONet} $\onet$ is a composition of \glspl{MLP}:
\begin{equation}
    (x, \psi) \mapsto f_d(f_x(x)\odot f_\psi(\psi)),
    \label{eq:onet}
\end{equation}
where $\odot$ is element-wise multiplication, $f_x$ is an encoder for the spatial points $x$, $f_\psi$ is an encoder for the \gls{PDE} parameters $\psi$, and $f_d$ is a decoder for the predicted state variable. \glspl{MLP} have $\tanh$ activation functions. Each $\onet$ is trained by using \gls{SGD} and the Adam~\cite{Kingma2014adam} optimizer over tuples $(x_n, \psi^d, w^d_n)$ to minimize the squared errors $|w^d_{n,\ell} - \onet(x_n, \psi^d)|^2$.

Like the \gls{ONet}, for the \gls{NBF}, a state variable's predictive model $\nbf$ is based on a set of \glspl{MLP}:
\begin{equation}
    (x, \psi) \mapsto \sum_{j=1}^{N_{bf}} C_{j,\ell}(\psi) \phi_{j,\ell}(x),
    \label{eq:nbf}
\end{equation}
where $\phi_{j,\ell}$ and $C_{j,\ell}$ are \glspl{MLP}. The $\phi_{j,\ell}$ are basis functions and the $C_{j,\ell}$ are ``unknowns'' functions.

The basis functions $\phi_{j,\ell}$ are trained by first constructing a basis from the data. A state variable's training data are preprocessed by concatenating each vector $[w^d_{1,\ell},\hdots,w^d_{N,\ell}]$ into an $N\times D$ matrix $\mathcal{W}_\ell$ and then using the \gls{SVD} to calculate the vectors $z_{j,\ell} \in \mathbb{R}^N$ of $\mathcal{W}_\ell$'s row-space, where $m=1,\hdots,N_{bf}$ indexes over the basis elements. Then the basis functions $\phi_{m,\ell}$ are trained with \gls{SGD}, Adam~\cite{Kingma2014adam}, and step-based learning rate decay, over tuples $(x_n, w_{m,\ell,n})$ to minimize the squared errors $|\phi_{m,\ell}(x_n) - z_{m,\ell,n}|^2$.


After the basis functions, $\phi_{j,\ell}$, are fit, the $C_{j,\ell}$ functions are fit: 
\begin{equation}
    \left|\sum_{j=1}^{N_{bf}} C_{j,\ell}(\psi^d) \phi_{j,\ell}(x_n) - w^d_{n,\ell}\right|^2.
\end{equation}

\section{Results}\label{sec:results}

\subsection{Implementation details}

The primary difference between the \gls{NBF} implemented in this paper and ~\cite{Witman2022nbf} is that the \gls{NBF} model here uses of logarithmic-based normalization to predict the variable $\rho$ and does not use the Partial Differential Equations (PDE) as part of its fitted loss.

We assess the effectiveness of the neural basis assumption 
by working in a low-data setting. Thus, we sample ten parameter configurations to use as training data and evaluate on the remaining 431 configurations. Further work could use larger training sets, although, due to the high resolution of this data, training becomes computationally expensive in a larger-data setting. Models are implemented and trained in PyTorch~\cite{Paszke2019pytorch}, in double precision.
We use a basis of $N_{bf} = 10$ basis functions. Domain geometries, for sampling from interiors and boundaries, used the \texttt{geometry} module from DeepXDE~\cite{Lu2021deepxde}. Prior to training, density values $\rho$ were transformed using the natural logarithm. Then, all state variables were normalized, using the mean and standard deviation of the training data. 
Network hyperparameters are in~\Cref{sec:hyperparameters}.

\subsection{Model assessment}\label{sec:assessment}

We show results for the overall data fit for the evaluation data in~\Cref{tab:eval_results}. We compute the relative error for a given data field, $l$, and set of parameters, $d$, as
\begin{equation}
    \sqrt{\frac{\sum_n \left|C_{j,\ell}(\psi^d) \phi_{j,\ell}(x_n) - w^d_{n,\ell}\right|^2}{ \left|\sum_n w^d_{n,\ell}\right|^2}}.
\end{equation}
The relative error results in~\Cref{tab:eval_results} are averaged over the set of test parameters, $l$. The \gls{NBF} approach is able to predict $u_1$, $u_2$, and $T$ well across the test set, despite having a low amount of training data. In particular, the \gls{NBF} performed significantly better than the \gls{ONet} baseline for the velocity and temperature predictions. This shows the benefit of learning a neural basis representation of the data. However, the error in predicting $\rho$ was poor for both models. This performance may be due to the order-of-magnitude variation in density values for different equation parameters, which the use of a log transform was not sufficient to mitigate. Potentially because predicting $\rho$ was so challenging, adding physics-informed regularization based on the continuity and momentum-conservation equations (Eqs.~\ref{eq:continuity} and~\ref{eq:momentum} in~\Cref{sec:governing_equations}) did not improve accuracy. Other work has shown the difficulties in training added by imposing physics-informed regularization~\cite{wang2021understanding,krishnapriyan2021characterizing,New2023pinns}.

\begin{table}[ht]
    \centering
    \begin{tabular}{l||l|l|l|l}
    Model                 & $u_1$ error (std)       & $u_2$ error (std) & $\rho$ error (std)& $T$ error (std)\\\hline\hline
    DeepONet              &     0.154 (0.118)              &    0.758 (1.09)      &     0.729 (0.170)         &  0.515 (0.242)         \\
    NBF                 &     0.0863 (0.0518)              &    0.148 (0.0960)      &     0.909 (0.448)         &  0.174 (0.114)         \\
    \end{tabular}
    \caption{Model evaluation: We evaluate our models with each state variable's relative error, averaged over all the configurations in the test split.
    }
    \label{tab:eval_results}
\end{table}

\begin{figure}
\begin{center}
\subfloat[True Value]{\includegraphics[width=0.3\linewidth]{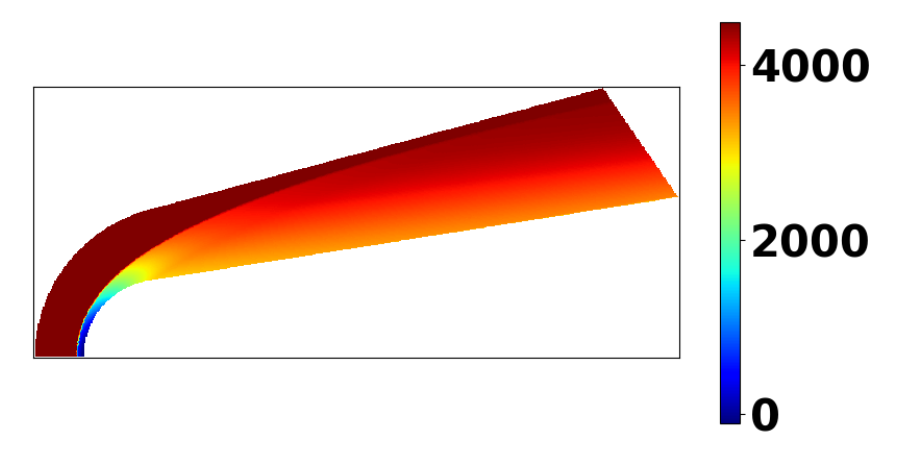}}
\hfill
\subfloat[Prediction]{\includegraphics[width=0.3\linewidth]{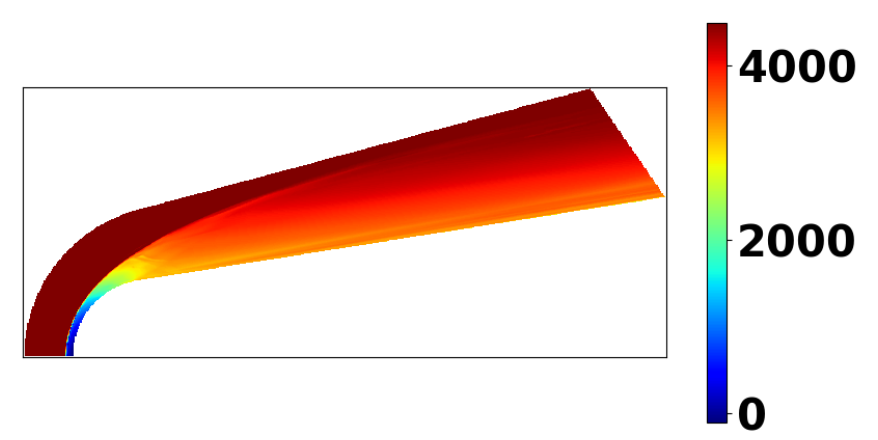}}
\hfill
\subfloat[Absolute Error]{\includegraphics[width=0.3\linewidth]{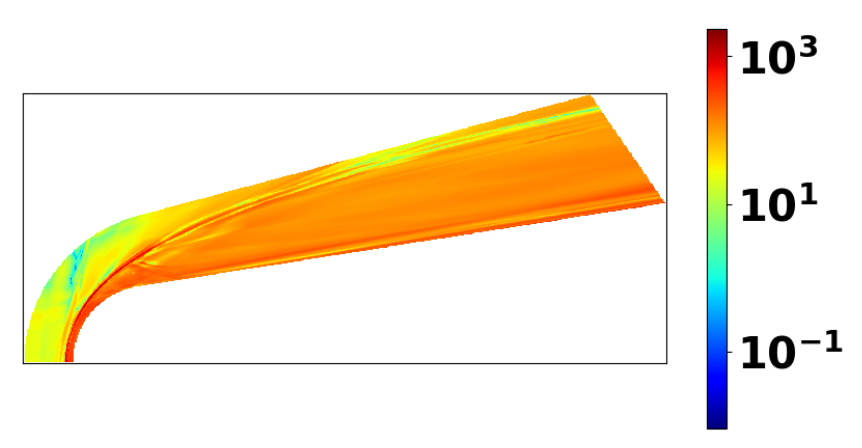}}
\caption{$x$-velocity $u_1$ at Mach $15$ and altitude \SI{26}{km} predicted by \gls{NBF} on evaluation data.}
\label{fig:x-vel}
\end{center}
\end{figure}

\begin{figure}
\begin{center}
\subfloat[True Value]{\includegraphics[width=0.3\linewidth]{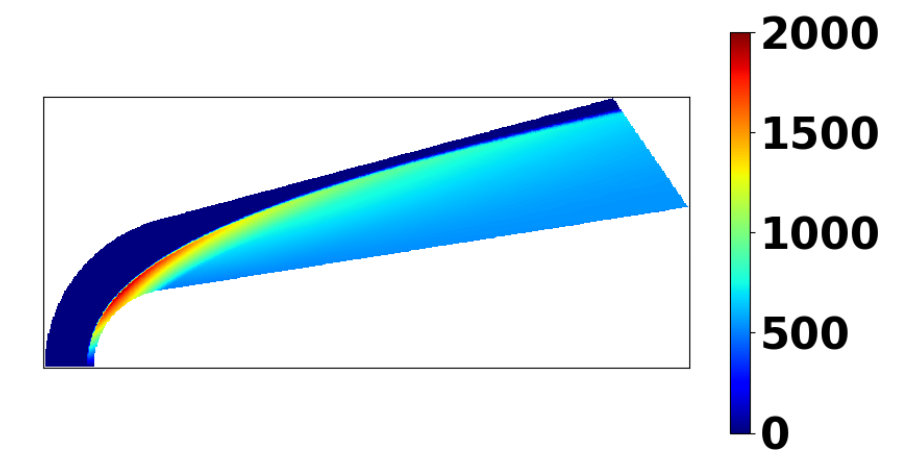}}
\hspace{.25cm}
\subfloat[Prediction]{\includegraphics[width=0.3\linewidth]{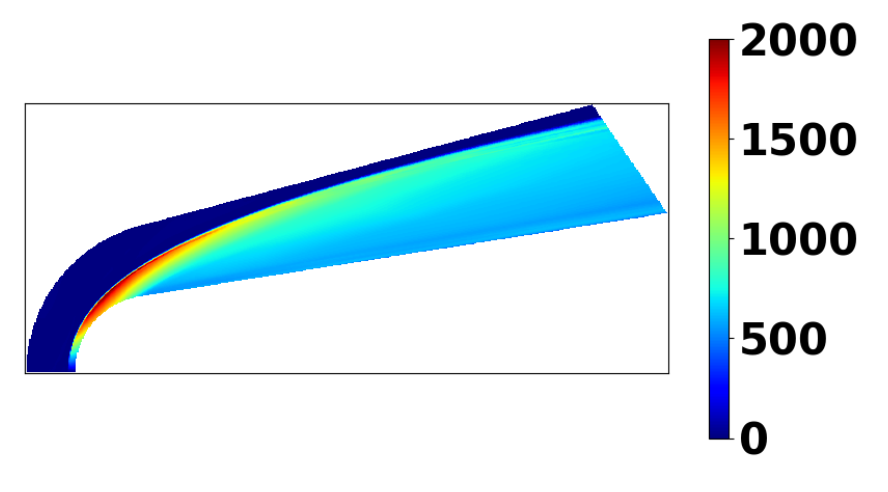}}
\hspace{.25cm}
\subfloat[Absolute Error]{\includegraphics[width=0.3\linewidth]{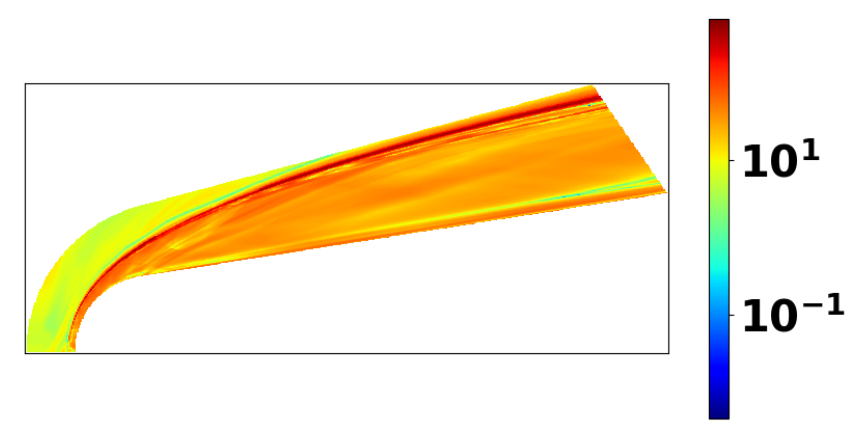}}
\caption{The $y$-velocity $u_2$ at Mach $15$ and altitude \SI{26}{km}, predicted by \gls{NBF} on evaluation data.}
\label{fig:y-vel}
\end{center}
\end{figure}

\begin{figure}
\begin{center}
\subfloat[True Value]{\includegraphics[width=0.3\linewidth]{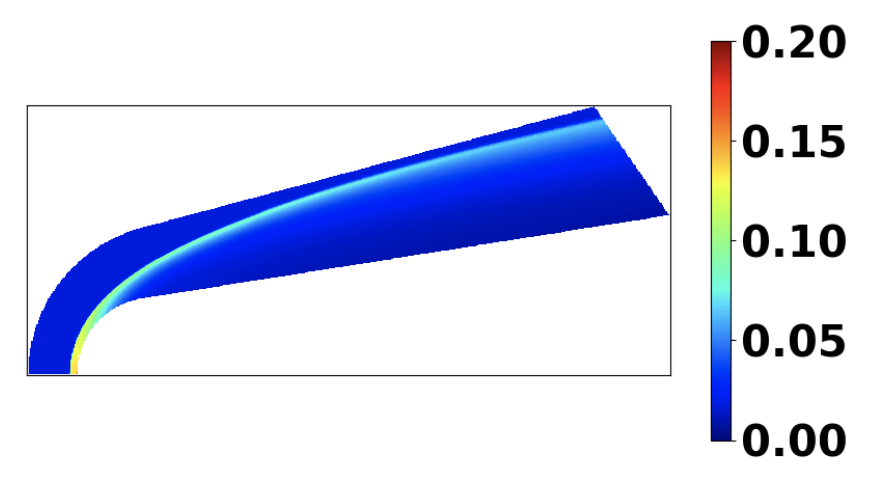}}
\hspace{.25cm}
\subfloat[Prediction]{\includegraphics[width=0.3\linewidth]{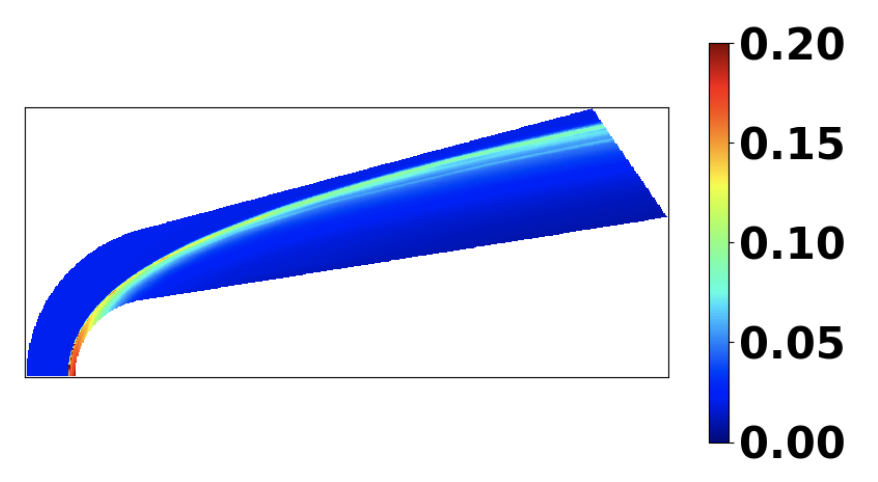}}
\hspace{.25cm}
\subfloat[Absolute Error]{\includegraphics[width=0.3\linewidth]{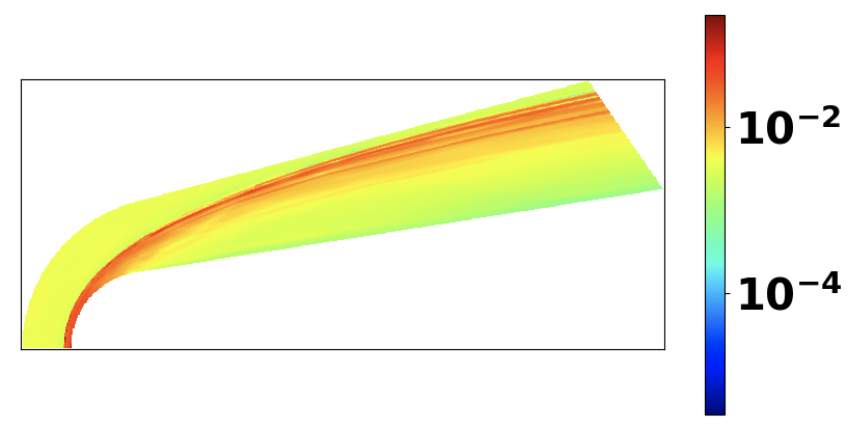}}
\caption{Density $\rho$ at Mach $15$ and altitude \SI{26}{km}, predicted by \gls{NBF} on evaluation data.}
\label{fig:rho}
\end{center}
\end{figure}

\begin{figure}
    \begin{center}
    \subfloat[True Value]{\includegraphics[width=0.3\linewidth]{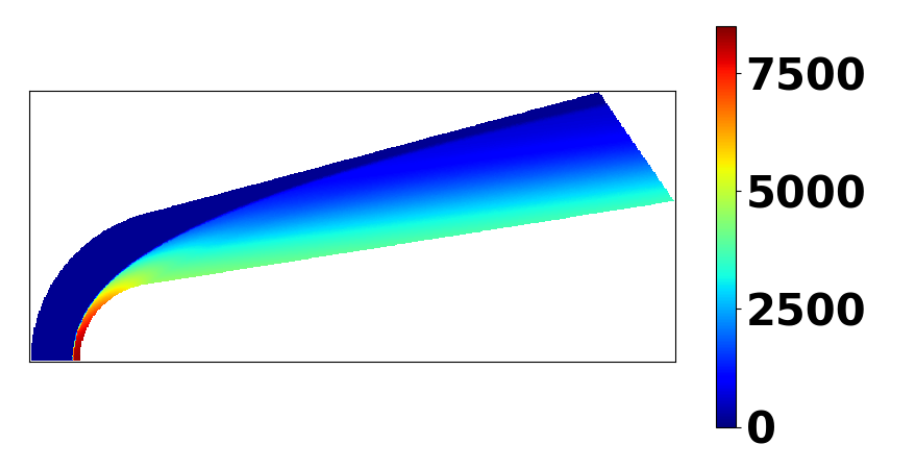}}
    \hfill
    \subfloat[Prediction]{\includegraphics[width=0.3\linewidth]{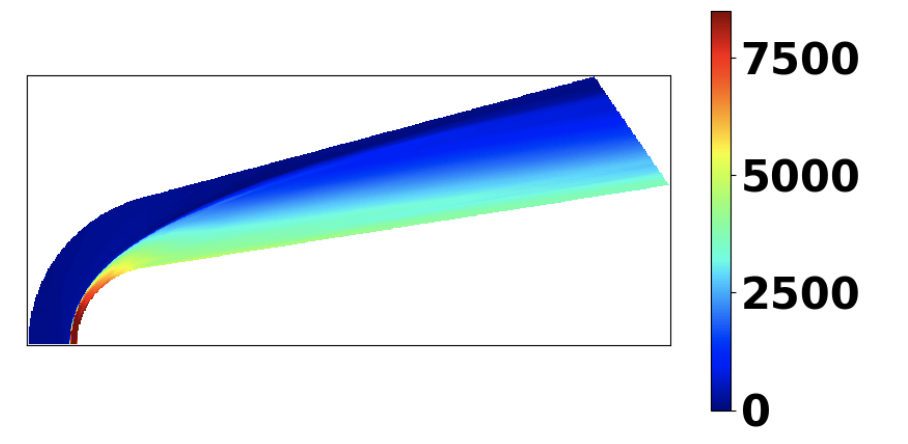}}
    \hfill
    \subfloat[Absolute Error]{\includegraphics[width=0.3\linewidth]{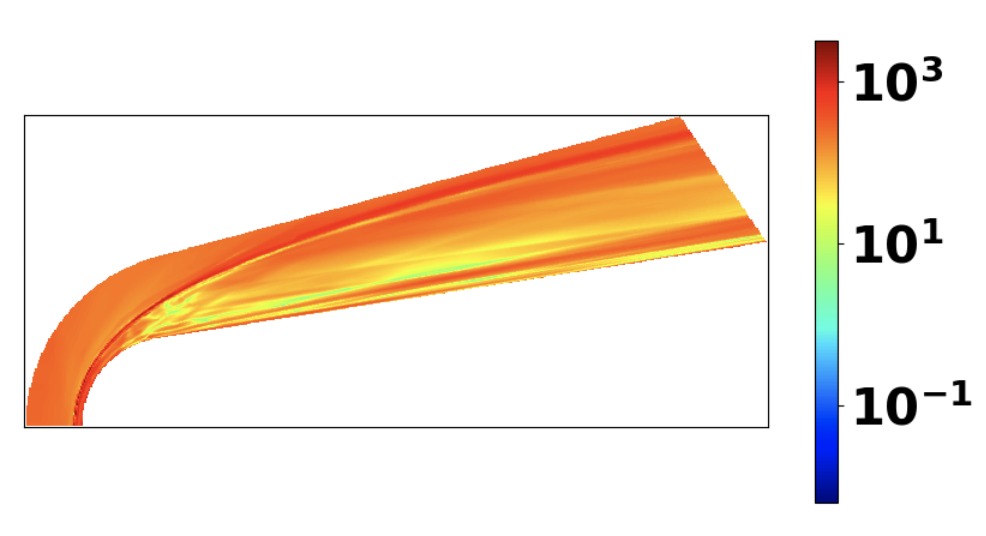}}
    \caption{Temperature $T$ at Mach $15$ and altitude \SI{26}{km}, predicted by \gls{NBF} on evaluation data.
    }
    \label{fig:temp}
    \end{center}
\end{figure}

In~\Cref{fig:x-vel,fig:y-vel,fig:rho,fig:temp}, the predictions are accurate for Mach $15$ and altitude of \SI{26}{km}. We observe that the \gls{NBF} has a tendency to have ridges of higher error, which we believe are a result of using a linear combination of bases with ridge-like data shapes.

It is also informative to assess how well the models fit the training data, which we show in~\Cref{tab:training_results}. The \gls{NBF} has significantly lower errors as compared to the \gls{ONet}. However, fitting the RAM-C II data remains challenging for both \gls{SciML} models, particularly for $\rho$. Even on the training set, no model surpasses relative error of $0.01$, and error for density $\rho$ remains high, at least $0.4$. Potential mitigation strategies include larger and more expressive models and more precise optimization strategies that include more than first-order derivative information (e.g.,~\cite{Byrd2016quasinewton,Yao2021_adahessian,Amid2022locoprop}).

\begin{table}[ht]
    \centering
    \begin{tabular}{l||l|l|l|l}
    Model                 & $u_1$ error (std)       & $u_2$ error (std) & $\rho$ error (std)& $T$ error (std)\\\hline\hline
    DeepONet              &     0.0567 (0.0162)              &    0.2154 (0.0284)      &     0.704 (0.159)         &  0.382 (0.0799)         \\
    NBF                 &     0.0361 (0.00217)              &    0.0803 (0.0184)      &     0.450 (0.138)         &  0.107 (0.0260)         \\
    \end{tabular}
    \caption{Model training fit: We report the data errors for our models with each state variable's relative error, averaged over all the configurations in the train split. 
}
    \label{tab:training_results}
\end{table}


\section{Conclusion}\label{sec:conclusion}
This paper demonstrates that a modified version of the NBF is able to accurately predict hypersonic fluid flow on a dataset of over 200 parameter values after training on just 10 data fields. The NBF method is able to predict the data fields in the validation set at a much higher accuracy than \gls{ONet}.

The NBF method can be used to model complex fluid flow in scenarios that only have a small amount of data. By improving the applicability of data-driven models for fluid flow, NBFs can be used to advance prediction, allowing for a faster engineering design process.

\section*{Acknowledgments}

This work was supported by internal research and development funding from the Air and Missile Defense Sector of the Johns Hopkins University Applied Physics Laboratory.

\bibliographystyle{unsrt}
\bibliography{references}

\appendix

\section{Data generation}\label{sec:data_generation}

We solve the compressible steady-state \gls{NSE} (\Cref{sec:governing_equations}) on a two-dimensional domain. Specifically, we simulate high Mach number flow over a blunt nose cone. This academic problem is based on a series of experiments that were performed in the 1960s: the \gls{RAM} flight experiments. These were designed to study plasma, specifically its effect on communication blackout during reentry. At high hypersonic Mach numbers, the temperatures surrounding the vehicle become sufficiently high that gas dissociates and ionizes, forming a thin plasma layer around the vehicle. The ionized air blocks out the incoming radio signal, causing communication blackout. 

The configuration of our study is based on the second flight test (RAM-C II). The RAM-C II vehicle can nominally be represented by a spherical blunt nose cone (\Cref{fig:ram-cII}a). The nose radius is \SI{0.1524}{m} and connects tangentially to the cone body, which has a half-cone angle of \SI{9}{^\circ}. The full body length of the configuration is \SI{1.3}{m}. 


Because the configuration is axisymmetric, we can simulate the system in two dimensions. \Cref{fig:ram-cII}b displays an example of the domain mesh along with its four boundaries (inflow, symmetry, cone wall and outflow). The inflow boundary is a Dirichlet \gls{BC}, and a Neumann \gls{BC} was used for the outflow. The symmetry line is a Neumman \gls{BC} for all its variables except for velocity component normal to the boundary; this component was forced to vanish at the boundary. The wall uses a mixed set of \glspl{BC}: the velocity is forced to vanish at the wall, the temperature is forced to equal \SI{1200}{K}, and the Neumann \gls{BC} is applied to the remaining dependent variables. For initializing the solver, the whole domain uses same condition as the inflow boundary.

To generate the training dataset, \gls{CFD} solutions were collected for various altitudes and Mach numbers.  The \gls{CFD} solver was used to solve the single species (air) \gls{NSE}. The solution inputs were parameterized by altitude and Mach number. Altitudes covered values of \SI{20}{km} to \SI{60}{km} in increments of \SI{2}{km}. Mach numbers covered values of $10$ to $30$ in increments of $1$. In total, $441$ solutions were generated. 

CFD++ version 20.1~\cite{web_cfd++} was utilized to perform all the simulations. CFD++ is an unstructured finite volume flow solver that is maintained by Metacomp Technologies Inc. It has the capability of solving the full Navier-Stokes equations with high-order spatial and temporal accuracy. To solve the single species Navier-Stokes equations, the solver uses a pseudo time marching approach. The initial condition is integrated over time until convergence is reached at which point we have our steady-state solution. Additional details on CFD++ can be found in the user manual~\cite{man_cfd++}. 

\section{Network hyperparameters}\label{sec:hyperparameters}

\Cref{tab:nbf_params} and~\Cref{tab:onet_params} give the hyperparameters used to train the \gls{NBF} and \gls{ONet} models. They were chosen based on experimentation with a similar RAM-C II dataset that used different configuration parameters and mesh points.

\begin{table}[ht]
    \centering
    \begin{tabular}{c|c}
    Hyperparameter                                                              &   Value\\\hline\hline
    \# of hidden units for each basis function $\phi_{j,\ell}$                  &   $40$\\
    \# of layers for each basis function $\phi_{j,\ell}$                        &   $7$\\
    \# of hidden units for each unknowns function $C_{j,\ell}$                  &   $100$\\
    \# of layers for each unknowns function  $C_{j,\ell}$                       &   $7$\\
    \# of hidden units for each nonlinear function $f_{\text{nonlin},\ell}$     &   $7$\\
    \# of layers for each nonlinear function $f_{\text{nonlin},\ell}$           &   $100$\\\hline
    \# epochs to train the basis functions                                      &   $100$\\
    learning rate for training the basis functions                              &   $10^{-3}$\\
    \# steps to decrease the learning rate for the basis functions              &   $45$\\
    learning rate reduction factor $\gamma$ for the basis functions             &   $0.9$\\\hline
    \# epochs to pretrain the \gls{NBF}                                         &   $10000$\\
    learning rate for training the basis functions                              &   $10^{-3}$\\
    \# steps to decrease the learning rate for pretraining                      &   $800$\\
    learning rate reduction factor $\gamma$ for pretraining                     &   $0.9$\\\hline
    \# epochs to train the \gls{NBF}                                            &   $1$\\
    learning rate for training the \gls{NBF}                                    &   $10^{-3}$\\
    \# steps to decrease the learning rate for the \gls{NBF}                    &   $8$\\
    learning rate reduction factor $\gamma$ for the \gls{NBF}                   &   $0.1$\\\\
    \end{tabular}
    \caption{Hyperparameters for training the \gls{NBF} models (Eq.~\ref{eq:nbf})}
    \label{tab:nbf_params}
\end{table}

\begin{table}[ht]
    \centering
    \begin{tabular}{c|c}
    Hyperparameter      &   Value\\\hline\hline 
    \# of hidden units for the spatial encoder $f_x$        &    $32$\\
    \# of layers for the spatial encoder $f_x$              &    $1$\\
    \# of hidden units for the parameter encode $f_\psi$    &    $32$\\
    \# of layers for the parameter encoder $f_\psi$         &    $1$\\
    \# of hidden units for the decoder $f_d$                &    $256$\\
    \# of layers for the decoder $f_d$                      &    $3$\\
    weight decay for training the \glspl{ONet}              &    $10^{-4}$\\
    \# epochs for training the \glspl{ONet}                 &    $97$\\\\
    \end{tabular}
    \caption{Hyperparameters for training the \gls{ONet} models (Eq.~\ref{eq:onet})}
    \label{tab:onet_params}
\end{table}

\section{Governing equations}\label{sec:governing_equations}

The four state variables of interest are $w = [u_1, u_2, \rho, T]$, where $u_1$ and $u_2$ are the velocities in the $x_1$ and $x_2$ directions, $\rho$ is the density, and $T$ is the temperature. The state variables obey the following compressible single-gas \gls{NSE}:

\begin{eqnarray}
    \sum_j \partial_{x_j}\left(\rho u_j\right) &=& 0\,\,\,\,\,(\text{continuity})\label{eq:continuity}\\
    \sum_j \partial_{x_j}\left(\rho u_i u_j + p \delta_{ij}\right) &=& \sum_j \partial_j \tau_{ij},\,\,\,\,i=1,2,\,\,\,(\text{momentum-conservation})\label{eq:momentum}\\
    \partial_j \partial_j \left[(E + p)u_j\right] &=& \sum_j\left[\sum_i \tau_{ij}u_i - k\partial_{x_j}T\right],\,\,\, (\text{energy-conservation)}.
\end{eqnarray}
Here, $E$ is the energy, given by $E = e + \frac{1}{2}\rho\sum_i u_i u_i$. The specific sensible enthalpy ($h$) satisfies the relations $h = e + R T$. The thermodynamic properties are modeled as polynomial-based function of temperature. For our case, 
\begin{eqnarray}
    h &=& R \left(a_{1}+\frac{a_{2}}{2}T+ \frac{a_{3}}{3}T^{2}+ \frac{a_{4}}{4}T^{3}+ \frac{a_{5}}{5}T^{4}\right) T + \Delta h_{f}^{0},
\end{eqnarray}
where the polynomial coefficients $a_1, a_2, a_3, a_4, a_5$ are determined empirically. For this study, we use a simplified model for air where $a_{1}\neq 0$ and $a_{2}=a_{3}=a_{4}=a_{5}= \Delta h_{f}^{0}=0$ ; thus, $h = R a_{1} T$. Viscous stress tensor ($\tau$) and strain rate tensor ($S$) are defined as following,

\begin{eqnarray}
    S_{ij} &=& \frac{1}{2}\left(\partial_{x_j} u_i + \partial_{x_i} u_j\right),\,\,\,\,i,j=1,2\\
    \tau_{jj} &=& 2\mu\left(S_{jj} - \frac{1}{3}\sum_i S_{ii}\right),\,\,\,\,j=1,2\\
    \tau_{ij} &=& 2\mu S_{ij},\,\,\,\,i\neq j,
\end{eqnarray}
and viscosity ($\mu$) and thermal conductivity ($k$) are defined by Sutherland's model:
\begin{eqnarray}
    \frac{\mu}{\mu_0} &=& \left(\frac{T}{T_{0,\mu}}\right)^{3/2}\frac{T_{0,\mu} + S_\mu}{T + S_\mu}\\
    \frac{k}{k_0} &=& \left(\frac{T}{T_{0,k}}\right)^{3/2}\frac{T_{0,k} + S_k}{T + S_k},
\end{eqnarray}
and $R$, $a_{1}$, $\mu_0$, $T_{0,\mu}$, $S_\mu$, $k_0$, $T_{0,k}$, and $S_k$ are constant parameters given in~\Cref{tab:ns_parameters}.


\begin{table}[ht]
    \centering
    \begin{tabular}{c|c|c}
        Parameter   &   Name                        &   Value\\\hline
        $R$         &   Specific gas constant       &   \SI{287}{\frac{J}{Kg\cdot K}}   \\
        $a_{1}$         &   Polynomial coefficient       &   $3.5$   \\
        $\mu_0$     &   Viscosity at reference temperature       &   \SI{1.716E-5}{\frac{kg}{m\cdot s}}   \\
        $T_{0,\mu}$ &   Reference temperature for viscosity       &   \SI{273.11}{K}   \\
        $S_\mu$     &   Sutherland constant  for viscosity    &   $111 K$   \\
        $k_0$       &   Thermal conductivity at reference temperature       &   \SI{2.41E-2}{\frac{W}{m\cdot K}}   \\
        $T_{0,k}$   &   Reference temperature for thermal conductivity       &   \SI{273.11}{K}   \\
        $S_k$       &   Sutherland constant for thermal conductivity       &   \SI{194}{K}   \\
    \end{tabular}
    \caption{\gls{NSE} parameters used to generate training and evaluation data}
    \label{tab:ns_parameters}
\end{table}

\end{document}